%% file: main.tex
\documentclass[10pt,twocolumn,letterpaper]{article}

\usepackage[pagenumbers]{cvpr} 

\usepackage{tikz}
\usepackage{multirow}
\usepackage{makecell}

\usepackage[T1]{fontenc}
\usepackage{sourcesanspro}       
\usepackage{fontaxes}            
\DeclareTextFontCommand{\textsb}{\fontseries{sb}\selectfont}


\definecolor{cvprblue}{rgb}{0.21,0.49,0.74}
\usepackage[pagebackref,breaklinks,colorlinks,allcolors=cvprblue]{hyperref}

\usepackage{algorithm}
\usepackage{algpseudocode} 

\def\paperID{1} 
\def\confName{CVPR}
\def\confYear{2026}

\title{ACCIDENT: A Benchmark Dataset for Vehicle Accident Detection \\ from Traffic Surveillance Videos}

\author{\textbf{Lukas Picek}$^{1,2,3}$, \textbf{Michal Čermák}$^{1}$ \textbf{Marek Hanzl}$^{1,3}$ and \textbf{Vojtěch Čermák}$^{1,4}$ \vspace{1mm} \\
$^1$PiVa AI, ~$^2$MIT, ~$^3$University of West Bohemia in Pilsen, ~$^4$Czech Technical University in Prague \\
{\tt\small lukaspicek@gmail.com / cermak.vojtech@seznam.cz / marek-hanzl@seznam.cz} \\
}
\definecolor{deepskyblue}{RGB}{0,191,255}

\begin{document}

\twocolumn[{%
\renewcommand\twocolumn[1][]{#1}%
\maketitle
\begin{center}
\vspace{-0.55cm}
    \includegraphics[width=0.975\linewidth]{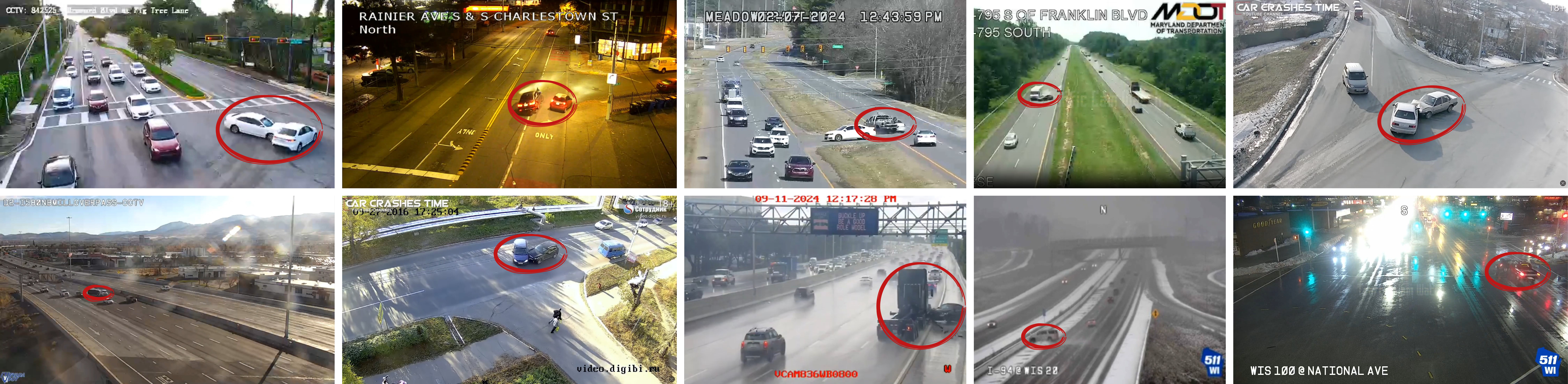}
    \captionof{figure}{\textbf{ACCIDENT dataset} reflects diversity of the real-world traffic footage. Available data vary in \textbf{scene type} (e.g., intersection, highway), \textbf{camera viewpoint}, \textbf{weather conditions} (e.g., rain, snow), \textbf{video quality} (e.g., resolution, compression), and \textbf{collision type} (e.g., head-on, rear-end, side-swipe), and originate from various regions across the globe (e.g., USA, Russia, the Middle East, Asia).}
    \label{fig:observation}
    \vspace{0.3cm}
\end{center}
}]

\begin{abstract}
We introduce \textsc{ACCIDENT}, a benchmark dataset for traffic accident detection in CCTV footage, designed to evaluate models in supervised (IID and OOD) and zero-shot settings, reflecting both data-rich and data-scarce scenarios. 
The benchmark consists of a curated set of 2,027 real and 2,211 synthetic clips annotated with the accident time, spatial location, and high-level collision type.
We define three core tasks: (i) temporal localization of the accident, (ii) its spatial localization, and (iii) collision type classification. Each task is evaluated using custom metrics that account for the uncertainty and ambiguity inherent in CCTV footage.
In addition to the benchmark, we provide a diverse set of baselines, including heuristic, motion-aware, and vision-language approaches, and show that ACCIDENT is challenging. You can access the ACCIDENT at: \url{https://accidentbench.github.io}

\end{abstract}

\input{sections/introduction}
\input{sections/related_work}
\input{sections/dataset}

\input{sections/benchmark}

\input{sections/method}
\input{sections/results}
\input{sections/conclusion}

\section*{Acknowledgment}
This research was supported by the Technology Agency of the Czech Republic, project No. TM05000009.

{
    \small
    \bibliographystyle{ieeenat_fullname}
    \bibliography{main}
}

\input{sections/appendix}

\end{document}

%% file: sections/introduction.tex
\vspace{-2mm}
\section{Introduction}
\label{sec:intro}

Road traffic accidents are a major global safety issue, resulting in over 1.2 million deaths annually \cite{world2023global}. They are the leading cause of death for people aged 5–29 and represent a growing concern, particularly in developing countries. Quick and reliable accident detection is critical since rapid emergency response can significantly improve survival and recovery outcomes \cite{rogers2015golden, batls2000}. Automated systems that can accurately and promptly detect accidents are therefore essential for improving response times and public safety.

The progress in accident detection has been primarily driven by first-person dashcam footage \cite{chan2016anticipating, Aditya_2021, fang2019dada2000drivingaccidentpredicted,Wang_2023_DeepAccident,fang2023vision}, which has enabled advances in anticipation by providing close-up views enriched with ego-motion cues. However, models trained on such data generalize poorly to surveillance footage. CCTV systems (already widely deployed) offer static, elevated views of multiple agents at once. This perspective introduces distinct challenges: videos are often of low resolution, affected by compression artifacts, occlusions, and poor lighting, and lack ego-motion cues that simplify accident detection in dashcam videos \cite{Xu2025tad}.

Although several traffic accident datasets exist, e.g., CADP~\cite{shah2018cadpnoveldatasetcctv}, TAD~\cite{Xu2025tad}, TADS~\cite{Chai2024tads}, and TUM~\cite{zimmer2025towards}, they differ in sensing setup, annotation granularity, and target task, and therefore provide only a fragmented view of accidents. Besides, they are often tailored to specific problem formulations, such as clipped accident detection, task-specific supervision, or multi-sensor roadside perception, which limits their utility as a unified benchmark for broader accident understanding in conventional CCTV footage. As a result, models trained on those datasets often capture only dataset-specific biases and transfer poorly across settings.

To address these limitations, we introduce ACCIDENT, a curated benchmark dataset\footnote{Any edited, duplicated, or otherwise unusable clips were removed. Labels were aggregated from annotations provided by 3--5 independent annotators using an uncertainty-aware fusion protocol.} for traffic accident detection in CCTV footage. It combines 2,027 real clips with 2,211 synthetic clips, with annotations for accident time, spatial location, and one of five high-level collision categories (\textit{head-on}, \textit{rear-end}, \textit{t-bone}, \textit{sideswipe}, or \textit{single-vehicle}); with the synthetic subset also providing bounding boxes, segmentation masks, and tracklets.

ACCIDENT is designed to support evaluation in fully supervised and zero-shot settings, thereby covering both data-rich and data-scarce scenarios. Across all settings, ACCIDENT defines three tasks: (i) temporal localization of the accident, (ii) spatial localization within the frame, and (iii) collision type classification, each evaluated with uncertainty-aware metrics that reflect the ambiguity of CCTV footage. To validate the benchmark, we provide a diverse set of baselines, including heuristic, motion-aware, and vision-language approaches, and show that ACCIDENT~is challenging across all settings.

\vspace{5pt}
\noindent\textbf{The core contributions of this paper include:}
\begin{enumerate}
\item A curated dataset of 2,027 real-world accident clips with temporal, spatial, and collision-type annotations.
\item A benchmark suite built for temporal and spatial localization, and collision-type classification, and evaluated under fully supervised, zero-shot, and sim-to-real settings using uncertainty-aware metrics.
\item A diverse set of baseline methods, including heuristic, motion-aware, and vision-language approaches.
\item A CARLA-based framework and 2,211 synthetic videos.

\end{enumerate}

%% file: sections/related_work.tex
\section{Related Work}
\label{sec:related_work}

Traffic accident understanding from video has received growing attention in anomaly detection, accident anticipation, and traffic-scene analysis~\cite{chan2016anticipating, sultani2019realworldanomalydetectionsurveillance, fang2019dada2000drivingaccidentpredicted, picek2025zero,zimmer2025towards}. However, much of the existing literature focuses on dashcam or ego-centric video, e.g., DAD~\cite{chan2016anticipating}, CCD~\cite{BaoMM2020}, YouTubeCrash~\cite{Aditya_2021}, A3D~\cite{yao2019unsupervisedtrafficaccidentdetection}, HTA~\cite{singh2020anomalousmotiondetectionhighway}, and DADA-2000~\cite{fang2019dada2000drivingaccidentpredicted}. Such datasets benefit from ego-motion, driver perspective, and in some cases auxiliary cues that are typically unavailable in fixed-view surveillance footage.

\begin{table}[!t]
\centering
\small
\begin{tabular}{@{}l|c|c@{\hspace{2.5mm}}c@{\hspace{2.5mm}}c@{\hspace{2.5mm}}c@{}}
\toprule
&  & \multicolumn{4}{c@{}}{\textbf{Type of annotations}} \\
\textbf{Dataset} & {\small \# clips}  & \textit{\small Object} & \textit{\small Time} & \textit{\small Location} & \textit{\small Type}   \\
\midrule
UCF-Crime~\cite{sultani2019realworldanomalydetectionsurveillance} & 150 & $\circ$  & $\circ$ & $\circ$  & $\circ$ \\
Smart City~\cite{tjtg-nz28-23} & 927 & $\circ$ & $\circ$ & $\circ$ & $\bullet$ \\
CADP~\cite{shah2018cadpnoveldatasetcctv} & \,\,240$^\dagger$ & $\bullet$ & {\color{gray}$\bullet$} & $\circ$ & $\circ$ \\
TAD~\cite{Xu2025tad} & 294 & $\circ$ & $\bullet$ & $\bullet$ & $\bullet$  \\
TADS~\cite{Chai2024tads} & \,\,966$^\ddagger$ & $\circ$ & $\bullet$ & $\bullet$ & {\color{gray}$\bullet$} \\
TUM Accid3nD~\cite{zimmer2025towards} & \,\,12 & $\bullet$ & $\bullet$ & $\bullet$ & $\bullet$ \\
\midrule
(our) ACCIDENT$^R$ & 2,027 & {\color{gray}$\bullet$} & $\bullet$ & $\bullet$ & $\bullet$    \\
(our) ACCIDENT$^S$ & 2,211 & $\bullet$ & $\bullet$ & $\bullet$ & $\bullet$    \\
\bottomrule
\end{tabular}
\caption{\textbf{Comparison of accident-related surveillance datasets.}
Existing datasets vary substantially in scale and annotation design, often providing only partial coverage of object, temporal, spatial, and collision-type labels. ACCIDENT unifies accident time, location, and type annotations across real ($R$) and synthetic ($S$) data. $^\dagger$Only a subset includes temporal labels. $^\ddagger$Classes not defined by collision geometry.}

\label{tab:category_counts}
\vspace{-3mm}
\end{table}

\medskip
\noindent\textbf{Surveillance-based accident datasets.}
Compared with dashcam datasets, existing surveillance-based accident datasets remain fragmented in sensing setup, scale, and annotation design (see Table~\ref{tab:category_counts}). CADP~\cite{shah2018cadpnoveldatasetcctv}, TAD~\cite{Xu2025tad}, and TADS~\cite{Chai2024tads} are largely built around short, pre-trimmed accident clips and task-specific annotations. 
TAD and TADS are sourced primarily from China. TUM Accid3nD~\cite{zimmer2025towards}, although valuable, is geographically limited to a single location and a few accidents, and it also involves complex sensing with multiple cameras and LiDARs.
CADP is more geographically diverse, but richer spatio-temporal annotations are available only for a subset of the data. 

As a result, prior datasets do not form a unified benchmark for jointly modeling \emph{when} an accident occurs, \emph{where} it happens, and \emph{what} type of collision is involved in conventional CCTV footage. This fragmentation makes cross-dataset comparison difficult and limits their use for broader accident understanding and, most importantly, deployment.

\begin{figure*}[!t]
    \centering
    \includegraphics[height=3.35cm]{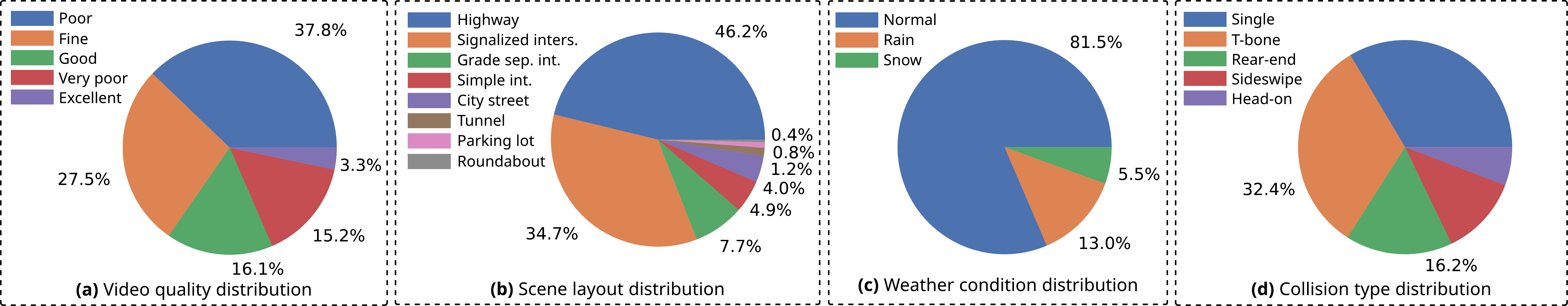}
\caption{\textbf{Diversity and challenge factors in the ACCIDENT dataset.} \textbf{(a)} Video quality is biased towards low and medium, reflecting the typical quality for CCTV feeds. \textbf{(b)} Scene layouts are dominated by highways and intersections with multi-agent scenarios, critical for robust model evaluation. \textbf{(c)} Weather distributions include a range of rain and snow conditions, enabling stress testing under adverse environments. \textbf{(d)} Accident types are well distributed, avoiding over-concentration in a single class and ensuring balanced evaluation.
}\vspace{-2mm}
    \label{fig:dataset-statistics}
\end{figure*}

\medskip
\noindent\textbf{Synthetic datasets} have been widely used to scale perception research in driving scenarios, e.g., CTAD~\cite{Luo2023CTAD}, DeepAccident~\cite{Wang_2023_DeepAccident}, IDDA~\cite{alberti2020idda}, OPV2V~\cite{xu2022opv2v}, SHIFT~\cite{sun2022shift}, and V2X-Sim~\cite{li2022v2xsimmultiagentcollaborativeperception}, but these resources primarily target ego-centric, cooperative, or multi-agent perception rather than CCTV-style accident understanding. Game-based datasets such as MP-RAD~\cite{Vijay2023MP-RAD} and GTACrash~\cite{Kim_Lee_Hwang_Suh_2019}, as well as other simulation tools~\cite{10313052}, are also less aligned with controlled benchmark generation for fixed-view surveillance settings.

\medskip
\noindent\textbf{Methods for accident detection.}
Prior methods span supervised classification, anomaly detection, and temporal event modeling. Early approaches often relied on RNNs, CNN-LSTM hybrids, or graph-based models, especially on dashcam datasets with strong ego-motion cues~\cite{chan2016anticipating,bao2021drive}. In surveillance settings, MIL-based classifiers, graph models, and offline CNN detectors have been explored~\cite{shah2018cadpnoveldatasetcctv,Chai2024tads}, while weakly supervised and unsupervised anomaly methods have used ConvLSTM~\cite{medel2016anomaly}, memory networks~\cite{gong2019memorizing}, and prediction--reconstruction strategies~\cite{liu2018future,Liu_2021_ICCV,al2024collaborative}. More recently, large vision-language models such as Qwen-VL~\cite{bai2025qwen2} and Video-LLaMA~\cite{zhang2025videollama}, as well as traffic-oriented systems such as SeeUnsafe~\cite{zhang2025language} and TrafficVLM~\cite{dinh2024trafficvlm}, have shown strong video reasoning capabilities. However, their use for benchmarked accident localization and collision understanding in CCTV footage remains underexplored.

%% file: sections/dataset.tex
\section{ACCIDENT Dataset}
\label{sec:dataset}

ACCIDENT is a benchmark dataset for traffic accident detection in CCTV footage, designed to support evaluation across multiple settings, including fully supervised (IID and OOD), zero-shot, and sim-to-real transfer. All settings share the same task formulation and evaluation protocol, and differ only in the \textit{supervision} and the corresponding data splits.

The benchmark consists of: (i) a curated collection of 2,027 real CCTV accident clips and (ii) a CARLA-based synthetic set with 2,211 clips. Since the real clips were collected from public sources, a small overlap with prior datasets was unavoidable. Based on our analysis, 14 videos also appear in TAD~\cite{Xu2025tad}, and 218 appear in CADP~\cite{shah2018cadpnoveldatasetcctv}\footnote{If relevant, all overlapping clips were put in training splits only.}.

Although U.S. sources dominate, about 20\% of clips originate from non-U.S. regions, including the EU, the UK, Australia, China, Abu Dhabi, and India. Clips were selected for clarity and relevance while preserving characteristic challenges of surveillance video, such as low resolution, occlusions, and wide fields of view. Each real clip is annotated with the accident time, spatial location, and collision type. The synthetic subset follows the same accident taxonomy and additionally provides dense annotations, including boxes, masks, and track-level metadata. Detailed dataset statistics are provided in Appendix~\ref{app:dataset-stats} and Figure~\ref{fig:dataset-statistics}; licensing and availability details are given in Appendix~\ref{app:licensing}.

\medskip
\noindent\textit{\textbf{Note:}} \textit{\textsc{ACCIDENT} focuses on temporal and spatial localization and collision-type classification; annotations for detection, segmentation, and tracking are not provided.}

\begin{figure*}[t]
    \centering
    \includegraphics[width=0.975\linewidth]{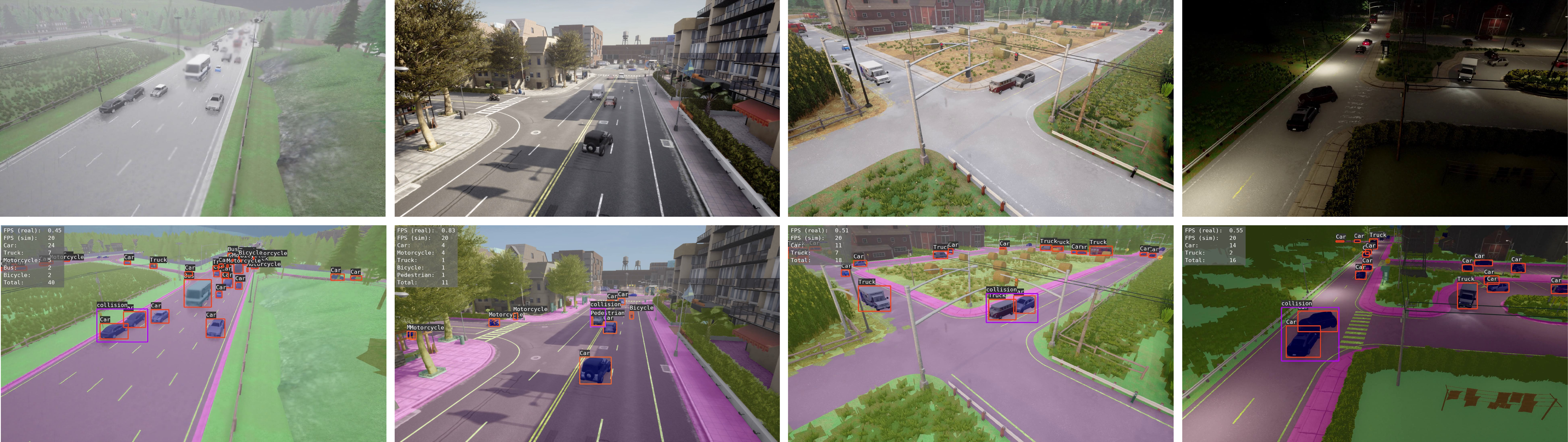}
\caption{\textbf{Synthetic accident scenarios.} Examples are generated from static third-person viewpoints that emulate CCTV. Road geometry, lighting, and weather are varied to stress-test detection under diverse conditions, while accident timing and agent behavior is controlled. \textbf{Top:} raw footage. \textbf{Bottom:} annotations showing class-labeled 2D boxes, collision indicator, and semantic segmentation.}
    \label{fig:dataset-carla}
    \vspace{-1mm}
\end{figure*}

\subsection{Data Preprocessing and Annotation}

\noindent\textbf{Data filtering and selection} included three main steps:
\begin{itemize}
    \item \emph{Screening:} Only fixed-view clips showing single- and multi-vehicle collisions were selected ( incidents involving pedestrians, cyclists, or animals were excluded for ethical reasons). Heavily edited videos (replays, overlays) and footage with severe compression artifacts, blur, or frame-skips were removed. \vspace{1mm}
    \item \emph{Cutting:} Because many source videos span hours of continuous traffic-camera footage, each clip was trimmed to a \(\sim\)30\,s segment around the impact frame. To reduce center bias, we applied a random offset to the temporal window. Slow-motion replays or drawn overlays (e.g., arrows or circles) were removed when they fell outside the segment; clips were discarded only if such edits overlapped the moment of the accident or obscured critical cues.\vspace{1mm}
    \item \emph{Frame deduplication:} duplicate or near-duplicate frames introduced by encoding glitches were pruned using an \texttt{\small ffmpeg}-based filter (static-frame removal/scene-change thresholding), without reordering frames. This step shortens some segments but preserves native FPS/resolution.
\end{itemize}

\medskip
\noindent\textbf{Annotation process} included labeling of (i) the \emph{impact frame} (first visible contact), (ii) a bounding box around collision participants, and (iii) a \emph{collision type} from a set of 5 classes (\texttt{\small head-on}, \texttt{\small rear-end}, \texttt{\small t-bone}, \texttt{\small sideswipe}, and \texttt{\small single-vehicle}) with \texttt{\small rollover} attribute marked when present. In addition, coarse scene/environment tags (e.g., intersection vs.\ highway; night/day; rain/snow) were recorded to enable stratified analysis.
To minimize subjectivity and capture ambiguity in low-resolution or occluded footage, each clip was annotated by 3–5 independent annotators. Disagreements were resolved by fixed consensus rules that produce a single label per target for evaluation: 
\textit{temporal} = median of annotated impact frames;
\textit{spatial} = a center \emph{point} calculated as the centroid of the box formed by per-axis means of annotated corners; 
\textit{collision type} = selected based on majority voting. In a rare case of a tie, an additional annotation is obtained until a majority is reached. For more info about the inter-annotator agreement, feel free to see the Appendix~\ref{app:iaa}.

\subsection{Synthetic Dataset}

In addition to the real CCTV clips, ACCIDENT includes a synthetic set designed to support benchmark settings with limited real-world supervision. The provided set contains 2,211 clips that cover all five collision types and follow the same high-level task formulation.

The clips were generated from fixed CCTV-like viewpoints across a range of road layouts, including highways, signalized intersections, roundabouts, etc. To increase diversity, each setup has been rendered repeatedly under varied weather, lighting, traffic, and camera configurations (see Figure~\ref{fig:dataset-carla} for examples). In addition, pre- and post-impact context is randomly trimmed so that the accident occurs at different times within the clip. Occasional near-misses and other unintended outcomes have been removed manually.

Unlike the real set, the synthetic data naturally provides more dense supervision. Each synthetic clip is therefore annotated not only with accident timing, spatial location, and collision type, but also with 2D bounding boxes, semantic segmentation, actor identities, trajectories, impact metadata, collision logs, and per-actor 3D state. We additionally provide synchronized RGB and LiDAR outputs, and export annotations in standard formats such as COCO and YOLO.

To generate this set, we build and provide a configurable CARLA-based framework. The framework enables systematic variation in road layout, environmental conditions, agent behavior, and collision types, and relies on editable configuration files to support reproducible, scalable generation. Implementation details, configuration templates, and sample scenarios are available on the \href{https://accident.github.io/}{ACCIDENT website}.

\section{Benchmarks and Data Splits}

ACCIDENT defines three benchmarks on the same underlying dataset, tasks, and evaluation metrics. The benchmark always addresses the same three questions (i.e., \textit{when}, \textit{where}, and \textit{what} kind of accident happened), only the train/test split and the available supervision are different.

These benchmarks capture different deployment scenarios. The \textit{in-distribution} tests performance in a controlled setting, the \textit{geo-aware} tests generalization under geographic shift, and the \textit{zero-shot} tests scenarios in which no labeled training data is available. See split statistics in Table~\ref{tab:stats}.

\noindent\textit{\textbf{Note:} For reference, we provide three splits, but any other custom split can be defined based on the use case. In all cases, clips known to overlap with TAD or CADP are kept in training only, and we recommend preserving this.}

\bigskip
\noindent\textbf{In-distribution split (IID)} acts like a standardized benchmark based on a stratified split of the real dataset. The partition is constructed so that key attributes, including collision type and major scene characteristics, follow as similar a distribution as possible across both sets, which minimizes unintended dataset shift.

\bigskip
\noindent\textbf{Geo-aware split (OOD)} aims to measure robustness under geographic shift. Its test set is restricted to the U.S., while the rest is available for training. At the same time, the split is constructed to remain comparable in size to the IID setting, so that performance differences are driven more by the geographic shift rather than by changes in data quantity.

\bigskip
\noindent\textbf{Zero-shot split} targets settings in which no labeled training data is available. It therefore contains no training partition, and methods are evaluated directly on the real dataset. This protocol is intended to reflect deployment scenarios where accident-specific supervision cannot be collected in advance and is particularly suited to prompt-based, instruction-based, and other training-free approaches.

\begin{table}[!b]
\vspace{-4mm}
\setlength{\tabcolsep}{8.0pt}
\centering
\begin{tabular}{@{}lc@{\hspace{3mm}}cc@{\hspace{3mm}}c@{}}
\toprule
\multirow{2}{*}{\textbf{Data Split}}  & \multicolumn{2}{c}{\textbf{Training}} & \multicolumn{2}{c}{\textbf{Test}} \\
 & \textit{Clips} & \textit{Fraction} & \textit{Clips} & \textit{Fraction} \\
\midrule
In-distribution & 507 & 25\% & 1,520 & 75\%  \\
Geo-aware       & 454 & 22.4\% & 1,573 & 77.6\%   \\
Zero-shot       & --  & 0\% & 2,027 & 100\% \\
\midrule
\textit{Synthetic} & 2,211 & 100\% & -- & 0\% \\
\bottomrule
\end{tabular}
\caption{\textbf{Benchmarks statistics.} The IID and OOD splits are sampled to maintain a similar training/test ratio.}
\label{tab:stats}
\end{table}

%% file: sections/benchmark.tex
\newpage
\section{Evaluation Metrics}
\label{sec:benchmark}

We evaluate performance across three core tasks: (i) temporal localization of the accident, (ii) spatial localization within the frame, and (iii) semantic classification of the collision type. Each task captures a different aspect of understanding accidents from CCTV footage, i.e., \textit{when} it happens, \textit{where} it happens, and \textit{why} it happens. We define a dedicated metric for each task, followed by a unified score that summarizes overall model performance.

\subsection{Temporal Localization of an Accident}

The goal of this task is to identify \textit{when} the accident occurs in a video. In CCTV footage, this moment is often difficult to determine precisely due to motion blur and occlusions. Therefore, we treat the problem as a \textit{peak prediction task}, where the goal is to indicate which \emph{time instant} (in seconds) is most likely to correspond to the accident.

We use a \textit{Gaussian similarity score} that softly penalizes deviations from the annotated accident time, which is robust to uncertainty and avoids the brittleness of hard thresholds. It assigns a score close to 1 when the predicted moment is near the ground-truth (GT) time, decaying smoothly with increasing temporal error.
Given a prediction in the form of a continuous confidence score $s(t)$ over time $t \in [0, L_i]$ for video $i$, where $L_i$ is the clip duration in seconds. The predicted accident time is defined as $\hat{t}_i = \arg\max_{t} s(t),$ and the per-video score as:
\begin{equation}
\mathcal{T}_i = \exp\!\left(-\frac{(t^*_i - \hat{t}_i)^2}{2\sigma_t^2}\right),
\label{eq:temporal_gaussian_seconds}
\end{equation}

where $t^*_i$ denotes the ground-truth accident time\footnote{GT is averaged over multiple annotators \emph{in seconds}. If a method outputs a frame index $\hat{f}_i$, we convert to time via $\hat{t}_i=\hat{f}_i/\mathrm{fps}_i$.} for video $i$, and $\sigma_t$ is a temporal tolerance parameter specified in seconds (e.g., $\sigma_t\in\{0.5,\,1,\,2\}$).

The final accident detection score over a dataset of $N$ videos is then defined as:
\begin{equation}
\mathcal{T} = \frac{1}{N} \sum_{i=1}^{N} \mathcal{T}_i
\end{equation}

\subsection{Spatial Localization of an Accident}

The goal of this task is to identify \textit{where} in the scene the accident occurs. Since the data involves multiple \textit{agents}, occlusions, and other factors, precise spatial localization becomes challenging. Rather than relying on accident detection, we reduce the problem to \textit{point localization}: representing the accident with an impact centroid coordinate ($\hat{x},\hat{y}$).

To evaluate predictions, we use an \textit{anisotropic Gaussian similarity score} that softly penalizes deviations from the annotated accident location in both horizontal and vertical directions. This formulation adapts to the shape of the annotated region and avoids binary thresholds, such as IoU.

Given a predicted point $(\hat{x}_i, \hat{y}_i)$ for video $i$, and the ground-truth box center $(x^*_i, y^*_i)$ with bounding box width $w_i$ and height $h_i$, the per-video score is defined as:
\begin{equation}
\mathcal{S}_i = \exp\left(-\left[ \frac{(x^*_i - \hat{x}_i)^2}{2\sigma_x^2} + \frac{(y^*_i - \hat{y}_i)^2}{2\sigma_y^2} \right] \right),
\end{equation}

where $\sigma_x$ and $\sigma_y$ 
reflect the spatial tolerance in the horizontal and vertical directions, respectively\footnote{We use mean norm. width ($\sigma_x$) and height ($\sigma_y$) of accident boxes.}. 
The final spatial localization score across $N$ videos is:
\begin{equation}
\mathcal{S} = \frac{1}{N} \sum_{i=1}^{N} \mathcal{S}_i.
\end{equation}

This metric corresponds to the exponential part of a 2D multivariate normal distribution with diagonal covariance and rewards predictions that fall close to the center of the annotated region, scaled relative to its size.

\subsection{Collision Type Classification}

The goal of this task is to classify the \textit{type} of the collision shown in the video. Each video is labeled with a single ground-truth category selected from a predefined set of collision types (e.g., rear-end, t-bone).

We evaluate performance using \textbf{Top-1 Accuracy}, which measures the percentage of videos for which the predicted class exactly matches the annotated label. Let $\hat{y}_i$ be the predicted class and $y^*_i$ the ground-truth label for video $i$. The final classification score over $N$ videos is defined as:
\begin{equation}
\mathcal{C} = \frac{1}{N} \sum_{i=1}^{N} \mathbb{1}[\hat{y}_i = y^*_i],
\end{equation}

where $\mathbb{1}[\cdot]$ is the indicator function, returning 1 if the prediction is correct and 0 otherwise.

\subsection{Unified Evaluation Score}

To enable a single-point comparison of model performance across all tasks, we define a unified evaluation metric that combines the three individual scores. Each component score is normalized to the range \([0, 1]\), with higher values indicating better performance:

\begin{itemize}[itemindent=5mm]
    \item $\mathcal{T}$ -- Temporal similarity score (Gaussian-based)
    \item $\mathcal{S}$ -- Spatial similarity score (Gaussian-based)
    \item $\mathcal{C}$ -- Type of accident classification accuracy (Top-1)
\end{itemize}

The unified score is computed as harmonic mean:
\begin{equation}
\text{ACC}^S 
= \frac{3}{\frac{1}{\mathcal{T}}+\frac{1}{\mathcal{S}}+\frac{1}{\mathcal{C}}}.
\end{equation}

This composite score encourages balanced solutions that perform well across all dimensions of the task, i.e., detecting when and where accidents occurred, as well as identifying why they happen. It also provides a single, interpretable value for leaderboard comparison and model selection.

%% file: sections/method.tex
\section{Baselines}
\label{sec:baselines}

To contextualize the challenging nature of ACCIDENT, we implement a wide variety of baselines for all three tasks: (i) temporal localization, (ii) spatial localization, and (iii) collision type classification. We provide five types of baselines: \emph{(a) naive} content-agnostic priors; \emph{(b) lightweight heuristics} (e.g., optical flow and bounding-box dynamics); (c)\,\emph{embedding-based classifiers} for collision type using generic visual features (e.g., DINOv2~\cite{oquab2023dinov2}, SigLIP2~\cite{tschannen2025siglip}); \emph{(d) zero-shot VLMs} (e.g., Qwen-VL2.5-7B~\cite{bai2025qwen2}, and Molmo-7B~\cite{deitke2025molmo}) used out of the box without task-specific tuning; and \emph{(e) human annotators} as an empirical upper bound and calibration check for our metrics. In addition, we also test \emph{sim-to-real transfer} by deriving class prototypes from synthetic data and evaluating on real clips.

\medskip
\noindent\textbf{Common setup.} We evaluate in two regimes: (i) \emph{oracle-aided} (non-target labels fixed to GT) and (ii) fully \emph{end-to-end} (sequential prediction). Metric definitions (Gaussian similarities for $\mathcal{T}$ and $\mathcal{S}$ with $\sigma_t$ and $(\sigma_x,\sigma_y)$; Top-1 accuracy for $\mathcal{C}$) are given in Sec.~\ref{sec:benchmark}.

\medskip
\noindent\textbf{Naive baselines.}
To establish a quantitative floor while accounting for time and location distributions, we propose a naive baseline for each task, that are deterministic and parameter-free. For temporal localization ($\mathcal{T}$), we use the clip’s midpoint in \emph{seconds}; for spatial localization ($\mathcal{S}$), we use the resolution-normalized image center $(0.5,0.5)$; and for collision type ($\mathcal{C}$), we select the majority class. 

\medskip
\noindent\textbf{Human upper bound.}
To quantify task difficulty, we use Inter-Annotator Agreement (IAA) as the human reference across all three tasks. IAA for each task is computed by scoring an individual’s annotations against the consensus label using the same metric as for other methods, and then averaging the scores across annotators. The per-task mean is reported as the human baseline. See App.~\ref{app:iaa} for full IAA protocol and additional analyses.

\medskip
\noindent\textbf{Object tracklets.}
In each video, we detect and track selected agent categories (e.g., \textit{car}, \textit{bus}, \textit{truck}) to obtain stable tracklets for the heuristic baselines. For detection, we use \texttt{YOLO11x}~\cite{yolo11_ultralytics} (conf: 0.15, NMS IoU: 0.7) with an input-scaling policy that sets the detector input to 640\,px when the longer image side is $\leq$750\,px and to 1280\,px otherwise; all boxes are mapped back to the native resolution. 
To produce per-object tracklets, we use \texttt{ByteTrack} (IoU match threshold 0.7, track buffer 30 frames). 

\begin{figure*}[t]
    \centering
    \includegraphics[width=0.965\linewidth]{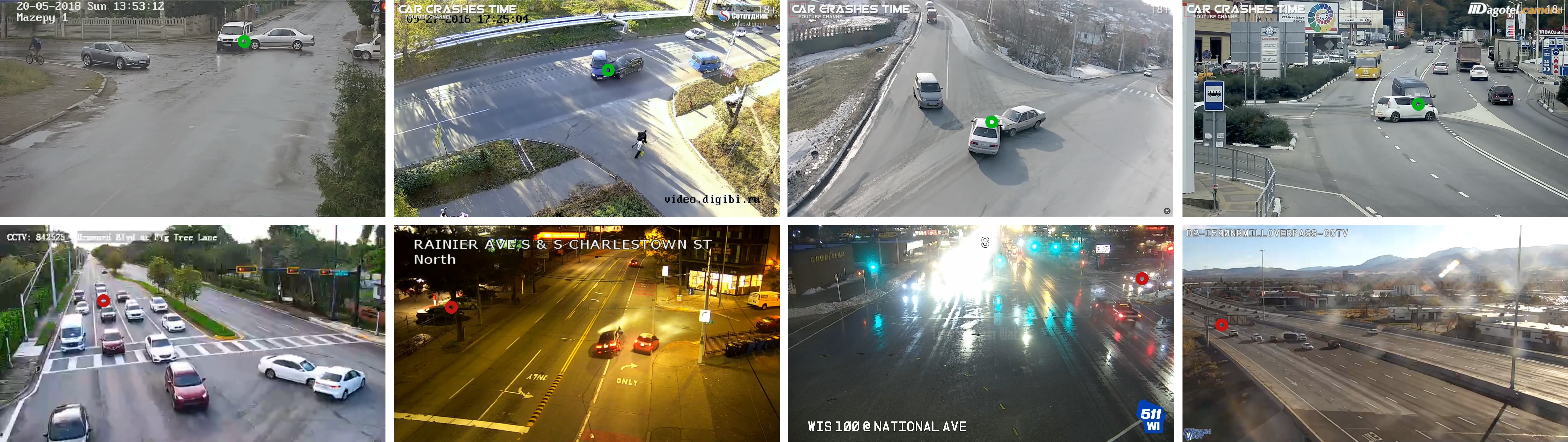}
\caption{
\textbf{Zero-shot spatial localization with Molmo-7B}. The top row shows representative successful localizations of the accident, while the bottom row shows typical failure cases, often caused by occlusion, poor lighting, low video quality, or ambiguous scenes.}
\vspace{-1mm}
\label{fig:qualitative}
\end{figure*}

\subsection{Temporal Localization}
\label{subsec:methods_T}

As baselines for temporal localization, we provide two training-free methods, \emph{Object Size Dynamics (OSD)} and \emph{Optical Flow (OF)}~\cite{picek2025zero}, their simple ensemble, and a zero-shot VLM methods (e.g., Qwen-VL2.5-7B, and Molmo-7B), which are reproducible, and label-agnostic by design.

\medskip
\noindent\textbf{Heuristic baselines}. Both proposed methods operate offline on a full-video time series produced either by OSD or OF. To suppress jitter and make the signals comparable, each series is smoothed with a 5-frame median filter and then standardised to zero mean and unit variance before change-point detection. For the ensemble, we take the accident onset as the simple average of the OSD and OF onsets (in seconds), which should reduce method-specific failure modes, i.e., missed detections in OSD or background-motion spikes in OF, and result in a more stable estimate. 

\begin{itemize}
\vspace{1mm}
    \item \emph{Object Size Dynamics.} Collisions typically induce abrupt, scene-agnostic changes in the geometry of tracked objects, i.e., boxes overlap, merge/split, or change scale due to contact and mutual occlusion, so the aggregate box area could serve as a direct proxy for impact that is robust to compression and low light. Building on the detected bounding boxes, we form a per-frame series \(A_t=\sum_k \mathrm{area}(\mathrm{box}_{k,t})\) and apply kernel change-point detection (KernelCPD)~\cite{ruptures-truong2020selective, celisse2018new, arlot2019kernel} to \(\{A_t\}\); the earliest change point defines the prediction. Aggregating across objects reduces sensitivity to brief ID switches or partial occlusions, and relying on box geometry keeps the method training-free, while still suitable for CCTV feed. \vspace{1mm}
    \item \emph{Optical Flow.} Impact events can be interpreted as sudden changes in the scene’s motion field, i.e., relative velocities spike at contact, and debris or vehicle rebound introduces transient motion bursts, making flow a complementary, class-agnostic cue. To capture this, we compute dense optical flow~\cite{farneback2003two} (using OpenCV~\cite{itseez2015opencv}) between consecutive frames and summarize each frame by a scalar motion series \(M_t=\tfrac{1}{HW}\sum_{x,y}\lVert \mathbf{u}_t(x,y)\rVert_2\), where \(\mathbf{u}_t\) is the 2D flow field and \(H{\times}W\) the frame size. Similar to OSD, we also apply KernelCPD to \(\{M_t\}\) and identify the earliest change point as the prediction.
\end{itemize}

\medskip
\noindent\textbf{Zero-shot VLM.}
To evaluate whether modern VLMs can separate pre- from post-impact frames in a zero-shot setting, we prompt the model at 0.5 second intervals with prompt:

\begin{quote}
\small\ttfamily
Is there a traffic accident or collision? Yes or No answer only.
\end{quote}

The answers are then collected over the frames and the first “Yes” defines the frame of an accident (converted to seconds for scoring). Both models are used out of the box with default inference settings and no task-specific tuning.

\subsection{Spatial Localization} 

Analogously to temporal localization, we use two zero-shot baselines for spatial localization: (i) an agent proximity and (ii) a VLM based, each operating on a single accident frame (GT in the oracle-aided regime, or the frame at the predicted onset \(\hat{t}\) in end-to-end) and producing a single 2D point in resolution-normalized coordinates \([0,1]^2\).

\medskip
\noindent\textbf{Bounding-box heuristic}.
At impact, the interacting agents are typically the closest pair in the scene, and their detection centers converge even under occlusion or scale changes. We therefore select the two objects with the smallest Euclidean distance between their centers and predict the midpoint between their centers. If only one valid detection is available, we use its center; if none, we fall back to the image center.

\medskip
\noindent\textbf{Zero-shot VLM.}
To get spatial predictions without task-specific tuning, we use the following prompt:
\begin{quote}
\vspace{-1.5mm}
\small\ttfamily
The scene depicts a traffic accident with one or more cars colliding. Point to the car accident.
\vspace{-1.5mm}
\end{quote}

Both models are used out of the box. If no point is returned, we take the image center. When multiple candidates are returned, we use the first one.

\subsection{Collision Type Classification}
\label{subsec:methods_C}

For collision-type classification, we evaluate two VLMs and transfer baselines based on pretrained DINOv2~\cite{oquab2023dinov2} and SigLIP2~\cite{tschannen2025siglip} embeddings with linear probes trained on the synthetic subset. For all experiments, we report results for two visual contexts: \emph{Full Image} and a \emph{Fixed Crop} centered on the ground-truth box, used as a reference upper bound. In the \emph{end-to-end} setting, the same classification procedure is applied to the frame at the predicted onset $\hat{t}$.

\medskip
\noindent\textbf{Zero-shot VLM.}
To get a single label prediction without task-specific tuning, we query the accident frame with:
\begin{quote}
\vspace{-1.5mm}
\small\ttfamily
What type of traffic accident is shown? Choose from: \{CLASS\_LABELS\} Return a single label.
\vspace{-1.5mm}
\end{quote}
Models are used out of the box with default inference settings; we normalize the response to the benchmark taxonomy using a keyword map. 

\medskip
\noindent\textbf{Linear Probing.}
For each image, we extract a feature vector using selected encoders (e.g., DINOv2 or SigLIP2). Then we train a linear classifier on top of the frozen embeddings using the subset of the synthetic dataset.

%% file: sections/results.tex
\section{Results}

\noindent\textbf{Temporal localization.}
Temporal localization is challenging across all baselines (Table~\ref{tab:temporal-baselines}). Overall, the heuristic methods provide the strongest performance, especially as the temporal tolerance is relaxed, while Molmo-7B is the most competitive VLM and performs best under the strictest tolerance. This suggests a trade-off between sharper but less consistent VLM predictions and more stable heuristic estimates. Although scores increase monotonically with larger $\sigma_t$, all automatic methods remain far below human performance (i.e., \(0.9794\pm0.0040\)), highlighting the difficulty of localizing accident onset in CCTV footage.

\medskip
\noindent\textbf{Spatial localization.}
Spatial localization seems substantially easier than temporal one, but is still far from solved (see qualitative examples in Figure~\ref{fig:qualitative}). Zero-shot VLMs clearly outperform the heuristic baselines, with Molmo-7B reaching \(\mathcal{S}=0.596\) and Qwen2.5-VL-7B \(0.436\), compared with \(0.273\) for the bounding-box heuristic and \(0.250\) for the content-agnostic prior. This indicates that ACCIDENT rewards semantic visual grounding more than simple spatial priors or hand-crafted proposals. At the same time, the gap to human performance remains large (\(0.9950\pm0.0015\)), leaving substantial room for improvement.

\begin{table}[!h]
\vspace{-1mm}
    \centering
    \setlength{\tabcolsep}{8pt} 
    \small
    \begin{tabular}{@{}l|ccc@{}}
    \toprule
    \textbf{Method} & $\sigma_{t=0.5}$ & $\sigma_{t=1}$ & $\sigma_{t=2}$ \\
    \midrule
    \textit{Naive Baseline}      & \textit{0.107} & \textit{0.190} & \textit{0.295} \\  
    \midrule
    Optical Flow  (OF)               & 0.140 & 0.266 & 0.466 \\ 
    Object Size Dynamics (OSD)       & 0.136 & 0.259 & 0.452 \\
    OF + OSD                         & \textbf{0.150} & \textbf{0.287} & \textbf{0.496} \\
    \midrule
        Qwen2.5-VL-7B    & 0.145 & 0.266 & 0.477 \\
        Molmo-7B         & \textsb{0.207} & \textsb{0.343} & \textsb{0.539}\\
    \midrule
    
    \textit{Human Performance}
        & \makecell{\textit{0.968}\\[-3pt]\tiny$\pm\textit{0.005}$}
        & \makecell{\textit{0.979}\\[-3pt]\tiny$\pm\textit{0.004}$}
        & \makecell{\textit{0.984}\\[-3pt]\tiny$\pm\textit{0.004}$} \\
    \bottomrule
    \end{tabular}
    \vspace{-2mm}
    \caption{\textbf{Zero-shot temporal localization.} Mean Gaussian similarity to impact time evaluated at three temporal tolerances ($\sigma{=}\{0.5,1,2\}$). \textit{Human performance} as mean$\;\pm\;$std.}
    \label{tab:temporal-baselines}
    \vspace{-1mm}
\end{table}

\newpage
\noindent\textbf{Collision Type Classification.}
Our baseline evaluation (see Table\,\ref{tab:category-baselines}) highlights a clear pattern: collision-type classification is almost impossible in the zero-shot regime, improves substantially with synthetic supervision, and remains sensitive to how much spatial context is provided. Both zero-shot VLMs perform poorly, with Molmo-7B consistently outperforming Qwen2.5-7B, but neither surpasses the majority-class baseline (0.335). This indicates that generic prompting alone does not provide enough structure for reliable collision understanding. In contrast, linear probes trained on synthetic data achieve much higher accuracy, showing that the task benefits from even lightweight supervision. The effect of cropping, however, is model-dependent: SigLIP2 improves when the view is restricted to the impact region (0.398 $\rightarrow$ 0.471), whereas DINO-v2 performs better on the full image (0.440 vs.\ 0.363), suggesting that local and global cues both matter, but not equally for all representations. Despite these gains, the best automatic results remain well below human performance (0.9228$_{\pm 0.0263}$).

\medskip
\noindent\textbf{End-to-end performance.}
Moving from oracle-aided evaluation to a fully chained pipeline exposes the unavoidable cost of error propagation. The temporal score \(\mathcal{T}\) is unchanged by construction, but downstream tasks become harder once they depend on imperfect upstream predictions. Spatial localization degrades when it must operate on the predicted onset \(\hat{t}\) rather than the ground-truth accident time, both for the heuristic pipeline (\(0.339 \rightarrow 0.273\)) and for Molmo-7B (\(0.596 \rightarrow 0.488\)). Collision-type classification is also affected: in the modular \textit{Best from all} setting, accuracy drops from \(0.440\) to \(0.433\). Notably, the strongest classifier differs between the isolated and end-to-end regimes: while SigLIP2 was slightly better in the earlier comparison, DINO-v2 proved more robust under predicted inputs and was therefore used in the final modular pipeline. Taken together, the tied results show that reliable accident understanding requires not only strong subtasks, but also robustness to imperfect temporal and spatial grounding.

\begin{table}[!h]
\vspace{-2mm}
\centering
\setlength{\tabcolsep}{7pt}
\small
\begin{tabular}{@{}ll|cc@{}}
\toprule

\multicolumn{2}{@{}l|}{\textbf{Method}}
& \textbf{Full image}
& \textbf{Fixed crop} \\
\midrule
\multicolumn{2}{@{}l|}{\textit{Naive Baseline}}  & \multicolumn{2}{c}{\textit{0.335}} \\
\midrule
\multirow{2}{*}{Zero-shot VLMs}
     & Qwen2.5-7B & 0.115 &  0.119 \\
     & Molmo-7B & 0.271 & 0.262  \\
\midrule
\multirow{2}{*}{Sim $\rightarrow$ Real} 
    & SigLIP2  & 0.398 & \textbf{0.471} \\
    & DINO-v2  & 0.440 & 0.363 \\
\midrule
\multicolumn{2}{@{}l|}{\textit{Human Performance}} 
& \multicolumn{2}{c}{\textit{0.9228}$_{\pm\text{\textit{0.0263}}}$} \\
\bottomrule
\end{tabular}
\caption{\textbf{Collision-type classification ($\mathcal{C}$) baselines.}
Top-1 accuracy across two visual contexts.
Zero-shot VLMs are evaluated without task-specific tuning. \emph{Sim$\rightarrow$real} classification follows the standard linear probing setting using the subset of synthetic data.
The \textit{Naive Baseline} is based on the majority class; \textit{Human Performance} reports mean$\pm$std across annotators.}
\label{tab:category-baselines}
\end{table}

\begin{table}[!b]
\vspace{-4mm}
\centering
\setlength{\tabcolsep}{3.2pt}
\small
\begin{tabular}{@{}l|ccc|ccc|c@{}}
\toprule
& \multicolumn{3}{c|}{\textbf{Oracle-Aided}} 
& \multicolumn{3}{c}{\textbf{End-to-End}} & \\
\textbf{Model} & $\mathcal{T}$ & $\mathcal{S}$ & $\mathcal{C}$ 
& $\mathcal{T}$ & $\mathcal{S}$ & $\mathcal{C}$ & ACC$^S$ \\
\midrule

\textit{Naive}
    & \textit{0.190} & \textit{0.250} & \textit{0.335}
    & \textit{0.190} & \textit{0.250} & \textit{0.335} & 0.245\\
\midrule
Heuristics  
    & 0.287 & 0.339 & -- & 0.287 & 0.273 & -- & -- \\
    
Molmo-7B
    & 0.343 & 0.596 & 0.271 & 0.343 & 0.488 & 0.293 & 0.358 \\
\textit{Best from all}
    & 0.343 & 0.596 & 0.471 & 0.343 & 0.488 & 0.433 & {\textbf{0.412}} \\

\midrule
\textit{Human}
    & \textit{0.979} & \textit{0.995} & \textit{0.923}
    & -- & -- & -- & -- \\
\bottomrule
\end{tabular}
\vspace{-1mm}
\caption{\textbf{Baseline end-to-end performance on ACCIDENT.} 
Temporal (\(\mathcal{T}\)) and spatial localization (\(\mathcal{S}\)), and collision-type classification (\(\mathcal{C}\)).
\emph{Oracle-aided} evaluates each component in isolation using ground-truth inputs, whereas \emph{End-to-End} chains predictions (\(\hat{t}\!\rightarrow\!\hat{p}\!\rightarrow\!\hat{c}\)) and therefore exposes error propagation across the pipeline.
\textit{Heuristics} uses OSD+OF for \(\mathcal{T}\) and the bounding-box midpoint for \(\mathcal{S}\).
\textit{Best from all} chains the strongest components from the preceding analyses; for \(\mathcal{C}\) we use DINO-v2 for its robustness to temporal and spatial errors.
It should therefore be interpreted as a modular upper bound rather than as a single model.
The \textit{Naive Baseline} is identical across regimes because it ignores visual content.
\textit{Human Performance} is shown for reference.}
\label{tab:oracle_vs_e2e}
\end{table}

%% file: sections/conclusion.tex
\section{Conclusion}
\label{sec:conclusion}

We introduced {ACCIDENT}, a benchmark for accident understanding in traffic-surveillance video that brings together three aspects of the problem that are usually studied separately: localizing \emph{when} the accident happens, detecting \emph{where} it occurs in the scene, and recognizing \emph{what} type of collision it is. Built from 2{,}027 curated real CCTV clips and complemented by 2{,}211 synthetic clips, the benchmark is designed to support evaluation across IID, OOD, zero-shot, and sim$\rightarrow$real settings under a common task definition and uncertainty-aware scoring protocol. In this way, {ACCIDENT} shifts the focus from dataset-specific shortcuts to the core difficulties of real surveillance video, where low quality, occlusion, and ambiguous event boundaries make accident understanding genuinely hard.

The baseline results show that {ACCIDENT} is challenging for reasons that are specific to real CCTV footage. Accident onset is difficult to pinpoint, spatial grounding is stronger but still imprecise, and collision-type recognition becomes fragile once timing and localization are imperfect. Even the strongest baselines are well below human performance, especially in end-to-end evaluation. Overall, the benchmark highlights that accident understanding in surveillance video requires joint reasoning about temporal change, spatial interaction, and collision semantics under severe visual uncertainty.

\medskip
\noindent\textbf{Outlook.} Progress will likely require models that couple change-point detection with spatial grounding under low-SNR CCTV conditions, reason over \emph{video} (not single frames), and exploit better domain adaptation methods. To better support future research, we establish an online leaderboard that is publicly available on the \href{https://accidentbench.github.io/}{Benchmark website}.

%% file: sections/appendix.tex
\newpage
\appendix

\section{Additional Dataset Statistics}
\label{app:dataset-stats}

The ACCIDENT includes 2,027 surveillance clips whose durations range from {1} to {114} seconds (median: 26.8s), having a median of {375} frames. Frame rates range from 4 to 50 FPS, reflecting the heterogeneity of online CCTV sources. Resolution ranges from 314p to 3840p, but visual clarity is often compromised by heavy compression, motion blur, and poor lighting. Therefore, around two-thirds are of poor or very poor quality.

Environmental diversity is relatively wide. Despite the majority of the videos being in normal weather (81.5\%), there are relatively large numbers of videos in rain (13\%) or snow (5.5\%), which add valuable edge cases. Nighttime scenes account for about a third of the videos, enabling evaluation under low-visibility conditions. Scene types are varied: almost half (46.2\%) occur at highways, with the remainder split across intersections, and other traffic layouts.

Accident types are distributed relatively evenly across the five defined structural types (e.g., head-on, sideswipe), ensuring broad task coverage and, importantly, reflecting the real distribution. Spatial annotations typically cover 0.7\% of the frame area, with most bounding regions falling within the {0.1--4.8\%} range. 

\section{Licensing \& Data Availability}
\label{app:licensing}
To ensure lawful and ethical use and reproducible evaluation, we define the licensing policy, describe the artifacts released, and establish privacy safeguards for \textsc{ACCIDENT}.

\medskip
\noindent\textbf{Policy \& provenance.}
Only clips whose upstream licenses \emph{explicitly permit redistribution and derivative works} (e.g., CC-BY, open-government reuse terms) were included; all others were excluded. For each released clip, the metadata stores a unique identifier and the canonical source URL.

\medskip
\noindent\textbf{Privacy.}
Given the overall small visual quality and the apparent size of scene agents, no clip contains visible faces; license plates may appear incidentally. No identity information is collected or annotated. Use is restricted to research, and any attempt at re-identification is prohibited. A takedown channel will be provided for removal requests.

\medskip
\noindent\textbf{Released artifacts \& licenses.}
We release: (i) downloadable videos for all licensed clips; (ii) per-clip annotations (impact time and location, and collision type) and metadata; (iii) the evaluation toolkit; and (iv) CARLA assets/code for development. Annotations/metadata are under \texttt{CC BY 4.0}; code and CARLA assets are under \texttt{Apache-2.0}.

\medskip
\noindent\textbf{Hosting.}
The ACCIDENT benchmark website, evaluation toolkit, code, and dataset access instructions are publicly available. The code is hosted on \href{https://github.com/accidentbench/ACCIDENT}{GitHub}, and the dataset is distributed through \href{https://www.kaggle.com/datasets/picekl/accident}{Kaggle}, and the leaderboards are available on the custom \href{https://accidentbench.github.io/}{Website}.

\section{Inter-Annotator Agreement (IAA)}
\label{app:iaa}

\noindent In order to quantify annotators’ uncertainty and the reliability of the ground truth, we provide an inter-annotator agreement (IAA) for all three tasks. These measurements establish an upper bound and allow us to better define $\sigma_t$, $\sigma_x$, and $\sigma_y$. Below, we report the per-annotator accuracy for six annotators\footnote{The A6 was the least active annotator with 700 annotated videos.}, and \textit{mean} for all. 
Besides, we provide confusion matrices for collision types for the best and worst annotators to expose systematic/semantically local confusions (see Fig.\,\ref{fig:iaa_confusions}).

\medskip
\noindent\textbf{Results.}
Table~\ref{tab:iaa_per_annotator} shows per-annotator accuracies for those with $>$700 labeled clips. Agreement is highest for \emph{spatial localization} (mean \(0.995\)), followed by \emph{impact frame} (mean \(0.979\)). \emph{Collision type} is lower on average (mean \(0.923\)) and exhibits larger cross-annotator variance (\(\pm 0.026\)), reflecting semantic ambiguity between certain classes (e.g., t-bone vs.\ sideswipe). These patterns are also clearly visible in the confusion matrices for A2 and A3 (Fig.~\ref{fig:iaa_confusions}).

\medskip
\noindent\textbf{Implications.}
The relatively larger spread for \emph{collision type} motivates clearer taxonomy definitions and examples in the guidelines. High \emph{spatial} and \emph{temporal}  agreement under our tolerance protocol supports the use of Gaussian similarity for evaluation (Sec.~\ref{sec:benchmark}), which softly penalizes small deviations while avoiding brittle thresholds.

\begin{table}[h]
\setlength{\tabcolsep}{4.25pt}
\small
\centering
\begin{tabular}{@{}lcccccc|c@{}}
\toprule
 & \textbf{A1} & \textbf{A2} & \textbf{A3} & \textbf{A4} & \textbf{A5} & \textbf{A6} &  {\scriptsize mean $\pm$\text{std}} \\
\midrule
$\mathcal{T}$ & 0.978 & 0.982 & 0.986 & 0.979 & 0.979 & 0.973 & $0.979  {\scriptsize \pm \text{0.004}}$ \\
$\mathcal{S}$ & 0.993 & 0.996 & 0.997 & 0.995 & 0.996 & 0.993 & $0.995 {\scriptsize \pm \text{0.001}}$ \\
$\mathcal{C}$ & 0.949 & 0.870 & 0.915 & 0.931 & 0.928 & 0.944 & $0.923 {\scriptsize \pm \text{0.026}}$ \\
\bottomrule
\end{tabular}
\caption{\textbf{Inter-annotator agreement accuracy.} 
Per-task accuracies against the consensus. Annotators with $>$700 clips only.
Rows denote $\mathcal{T}$=\emph{temporal}, $\mathcal{S}$=\emph{spatial}, $\mathcal{C}$=\emph{collision type}. For temporal localization, $\sigma_t{=}1$ is used.}
\label{tab:iaa_per_annotator}
\end{table}

\begin{figure}[!h]
\vspace{-3mm}
    \centering
    \setlength{\tabcolsep}{6pt} 
    \begin{tabular}{@{}c c@{}}
        {\hspace{12mm} \normalsize Annotator 1} & {\hspace{12mm} \normalsize Annotator 2} \\
        \includegraphics[width=0.475\linewidth]{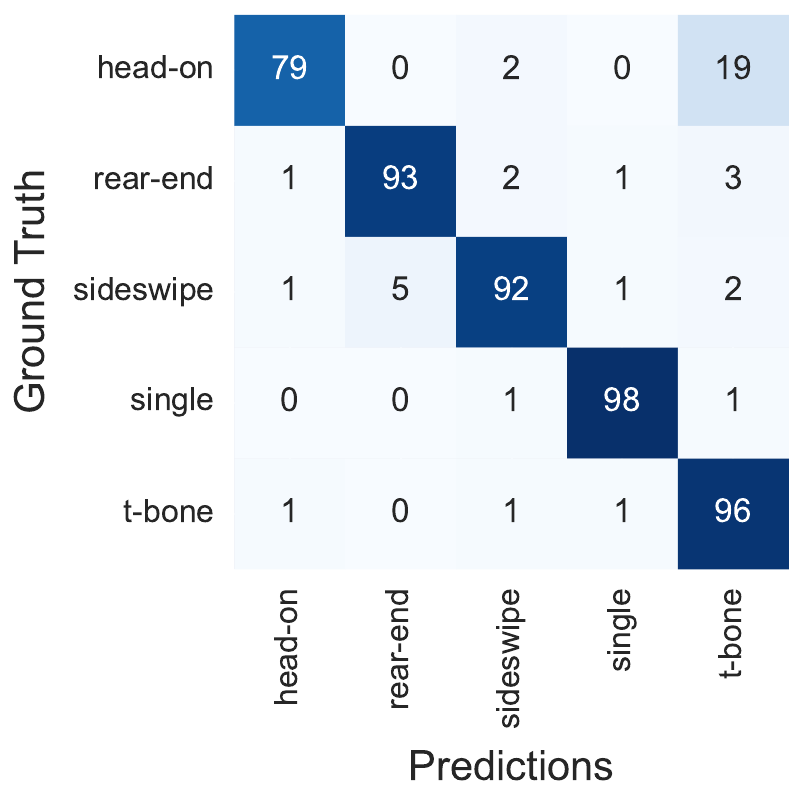} &
        \includegraphics[width=0.475\linewidth]{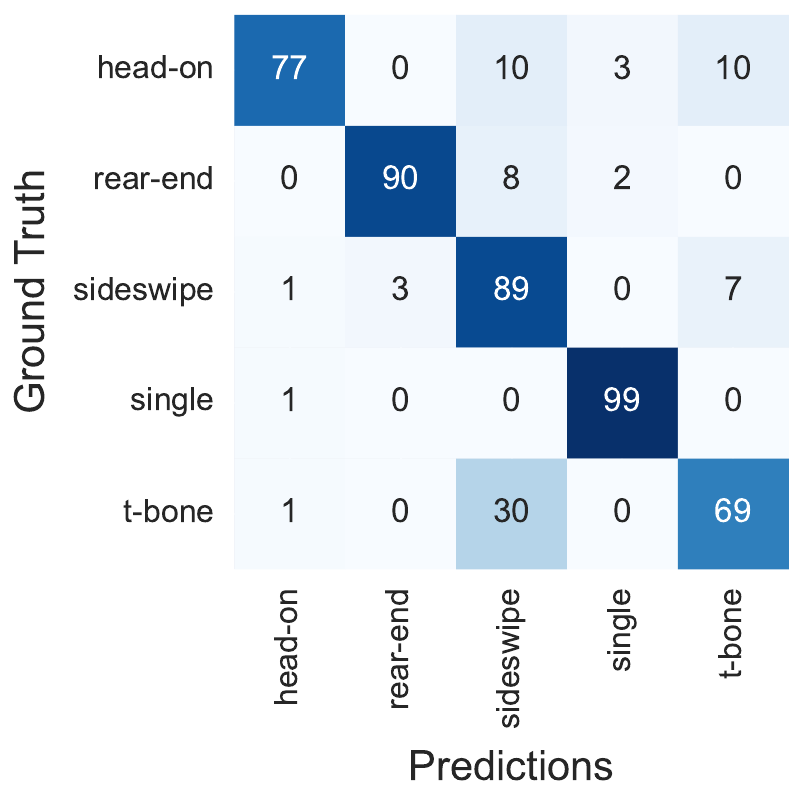}
    \end{tabular}
\caption{\textbf{Collision-type annotation confusion.} As expected, \textit{errors} concentrate between semantical similar classes (e.g., \emph{t-bone} vs.\ \emph{sideswipe}), with just handful of unrelated confusions.}
    \label{fig:iaa_confusions}
\end{figure}